\documentclass[sigconf]{acmart}

\AtBeginDocument{%
  \providecommand\BibTeX{{%
    \normalfont B\kern-0.5em{\scshape i\kern-0.25em b}\kern-0.8em\TeX}}}

\copyrightyear{2024}
\acmYear{2024}
\setcopyright{rightsretained}
\acmConference[KDD '24] {Proceedings of the 30th ACM SIGKDD Conference on Knowledge Discovery and Data Mining }{August 25--29, 2024}{Barcelona, Spain.}
\acmBooktitle{Proceedings of the 30th ACM SIGKDD Conference on Knowledge Discovery and Data Mining (KDD '24), August 25--29, 2024, Barcelona, Spain}
\acmISBN{979-8-4007-0490-1/24/08}
\acmDOI{10.1145/3637528.3672043}
\usepackage{makecell}
\usepackage{multirow}
\usepackage{subcaption}
\usepackage{bbding}
\usepackage{balance}
\begin{document}

\title{Learning Multi-view Molecular Representations with Structured and Unstructured Knowledge}

\author{Yizhen Luo}
\affiliation{%
  \institution{Institute of AI Industry Research (AIR), Tsinghua University}
  \state{Beijing}
  \country{China}
}
\email{yz-luo22@mails.tsinghua.edu.cn}

\author{Kai Yang}
\affiliation{%
  \institution{Institute of AI Industry Research (AIR), Tsinghua University}
  \state{Beijing}
  \country{China}
}
\email{yangkai@air.tsinghua.edu.cn}

\author{Massimo Hong}
\affiliation{%
  \institution{Institute of AI Industry Research (AIR), Tsinghua University}
  \state{Beijing}
  \country{China}
}
\email{hongcd21@mails.tsinghua.edu.cn}

\author{Xing Yi Liu}
\affiliation{%
  \institution{Institute of AI Industry Research (AIR), Tsinghua University}
  \state{Beijing}
  \country{China}
}
\email{liuxingyi99@gmail.com}

\author{Zikun Nie}
\affiliation{%
  \institution{Institute of AI Industry Research (AIR), Tsinghua University}
  \state{Beijing}
  \country{China}
}
\email{nzk20@mails.tsinghua.edu.cn}

\author{Hao Zhou}
\affiliation{%
  \institution{Institute of AI Industry Research (AIR), Tsinghua University}
  \state{Beijing}
  \country{China}
}
\email{zhouhao@air.tsinghua.edu.cn}

\author{Zaiqing Nie}
\authornote{Corresponding author.}
\affiliation{%
  \institution{Institute of AI Industry Research (AIR), Tsinghua University}
  \institution{Pharmolix Inc.}
  \state{Beijing}
  \country{China}
}
\email{zaiqing@air.tsinghua.edu.cn}

\renewcommand{\shortauthors}{Luo et al.}

\begin{abstract}
  Capturing molecular knowledge with representation learning approaches holds significant potential in vast scientific fields such as chemistry and life science. An effective and generalizable molecular representation is expected to capture the consensus and complementary molecular expertise from diverse views and perspectives. However, existing works fall short in learning multi-view molecular representations, due to challenges in explicitly incorporating view information and handling molecular knowledge from heterogeneous sources. To address these issues, we present MV-Mol, a molecular representation learning model that harvests multi-view molecular expertise from chemical structures, unstructured knowledge from biomedical texts, and structured knowledge from knowledge graphs. We utilize text prompts to model view information and design a fusion architecture to extract view-based molecular representations. We develop a two-stage pre-training procedure, exploiting heterogeneous data of varying quality and quantity. Through extensive experiments, we show that MV-Mol provides improved representations that substantially benefit molecular property prediction. Additionally, MV-Mol exhibits state-of-the-art performance in multi-modal comprehension of molecular structures and texts. Code and data are available at \url{https://github.com/PharMolix/OpenBioMed}.
\end{abstract}





\maketitle

\section{Introduction}
\begin{figure}[htpb]
\centering
\begin{subfigure}[b]{\linewidth}
    \includegraphics[width=\linewidth]{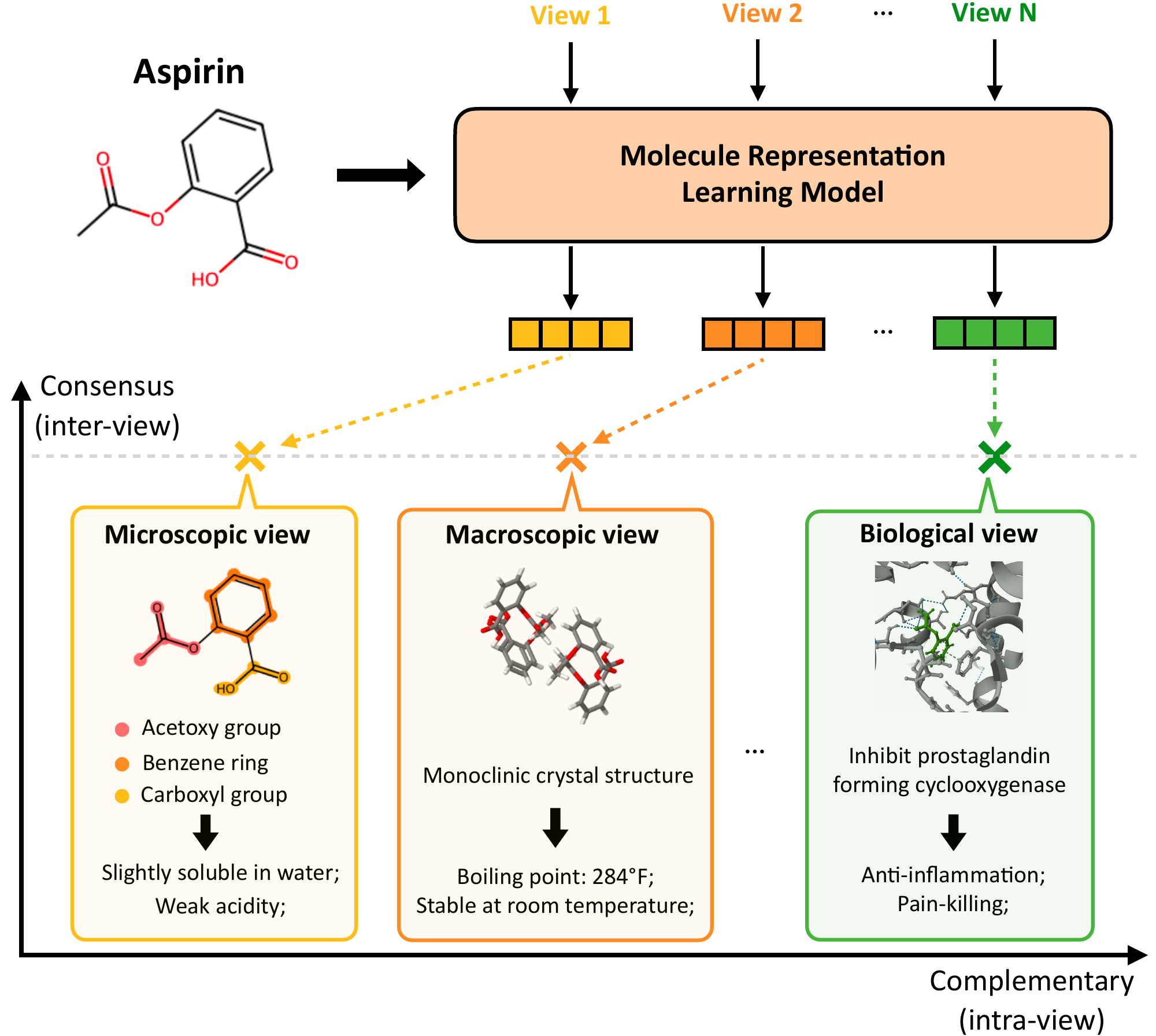}
    \caption{The multi-view characteristics of molecular expertise.}
    \label{fig:concept_mv}
\end{subfigure}
\begin{subfigure}[b]{\linewidth}
    \includegraphics[width=\linewidth]{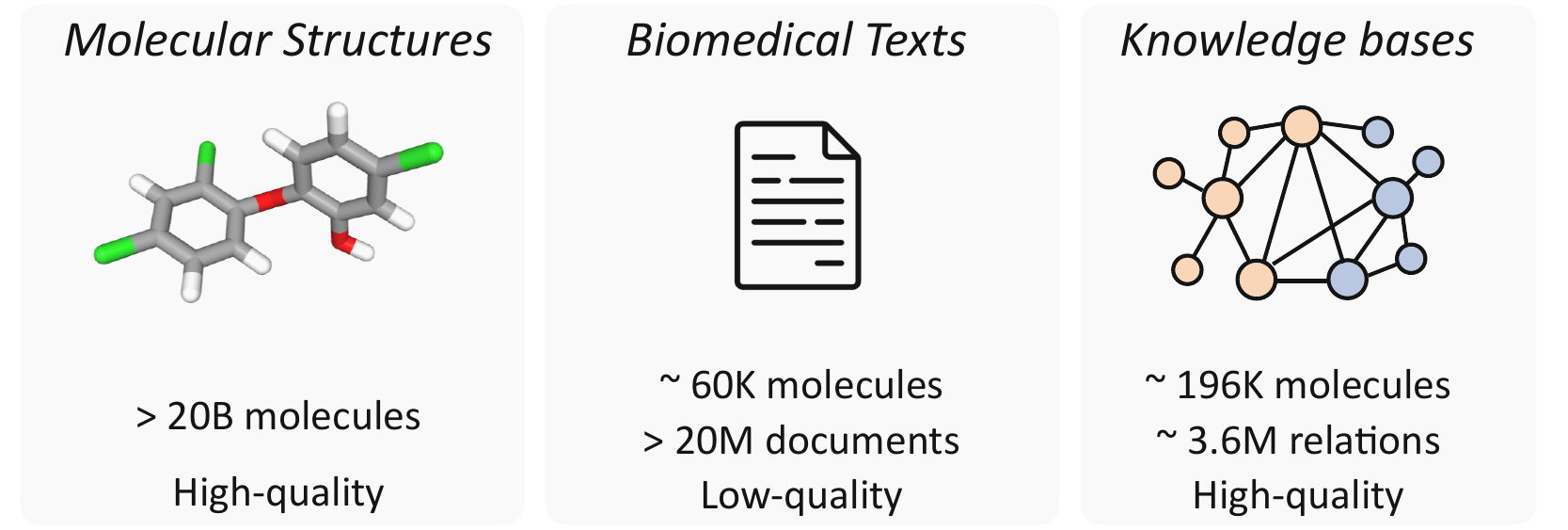}
    \caption{The heterogeneous sources of molecular expertise.}
    \label{fig:concept_mm}
\end{subfigure}
\captionsetup{font={small,stretch=0.95}}
\caption{An overview of molecular expertise. Molecular expertise covers diverse views that share consensus and complementary information. It resides within heterogeneous sources that vary in quality and quantity.}
\end{figure}
Understanding the properties and functions of small molecules is a pivotal issue in various scientific applications such as chemistry \cite{becker2001computational}, biology \cite{sliwoski2014computational}, and material design \cite{guo2021artificial}. Due to the substantial annotation costs for molecules, molecular representation learning (MRL), which aims at capturing molecular knowledge from abundant unlabeled data, has attracted significant research attention. Through self-supervised learning \cite{liu2021self} on various types of molecular structures, such as 1D SMILES strings \cite{chithrananda2020chemberta, irwin2022chemformer}, 2D graphs \cite{hu2020strategies, you2020graph, wang2022molecular} or 3D conformations \cite{zhu2022unified, stark20223d, zhou2023uni}, MRL models have achieved great success in advancing our understanding of molecules.

However, to further extend the application scope of MRL models, one is faced with a critical problem: molecular expertise is complicated and multifaceted, spanning diverse disciplines and views. It comprises consensus information shared by multiple views, as well as complementary information within each specific view \cite{chao2016consensus}. As exemplified in Figure \ref{fig:concept_mv}, the characteristics of Aspirin are identified through various perspectives. From a microscopic view, the molecule comprises a benzene ring, an acetoxy group, and a carboxyl group, resulting in slight water solubility and weak acidity. From a macroscopic view, it exhibits a monoclinic crystal structure, influencing physical properties such as boiling point and stability. From a biological view, it inhibits the production of prostaglandins by targeting cyclooxygenase, thus exhibiting anti-inflammatory and pain-killing functions \cite{ugurlucan2012aspirin}. A versatile MRL model is expected to generate view-based representations to address the distinctions between different application contexts. Unfortunately, existing works primarily focus on learning a universal representation to capture the consensus information across various views, and fall short in grasping the complementary characteristics of each view.


More recently, the emergence of MRL methods incorporating molecular knowledge from heterogeneous sources has opened novel avenues for addressing the multi-view problem. These models focus on jointly comprehending molecular structures, structured knowledge from knowledge bases, and unstructured knowledge from biomedical texts. The integration of heterogeneous inputs is accomplished either through language modeling objectives within a unified model \cite{zeng2022deep, liu2023molxpt, pei2023biot5}, or by contrastive learning objectives with independent encoders \cite{su2022molecular, liu2023multi, seidl2023enhancing}. Innovated by the success of these models, we aim to capture multi-view molecular expertise from these structured and unstructured knowledge sources. 

Nevertheless, there exist two challenges in learning multi-view molecular representations. First, as illustrated in Figure \ref{fig:concept_mv}, the MRL model should incorporate view information explicitly into its representations to ensure its adaptability in broad applications. However, prior MRL models integrate view information implicitly with 'wrapped texts' \cite{liu2023molxpt, pei2023biot5} or fine-tuning on downstream tasks, which compromises their understanding of the consensus and complementary relationships between molecular knowledge from different views. Besides, as shown in Figure \ref{fig:concept_mm}, it is essential to address the heterogeneity of information sources, including molecular structures, biomedical texts, and knowledge graphs, which vary in quality and quantity. Previous works treat structured and unstructured knowledge indiscriminately by transforming knowledge graphs into texts. However, this may introduce biases across different views due to the imbalanced distribution of pre-training data.

In this work, we propose MV-Mol, a comprehensive framework for \textbf{M}ulti-\textbf{V}iew \textbf{Mol}ecular representation learning with structured and unstructured knowledge, to address the aforementioned problems. To explicitly incorporate view information, we utilize text prompts to capture the complementary and consensus characteristics of different views for molecules. We leverage Q-Former \cite{li2023blip}, a multi-modal fusion architecture, to extract view-based molecular representations by jointly comprehending molecular structures and view prompts. Then, we propose a two-stage pre-training strategy to address the heterogeneity of structured and unstructured knowledge. The first stage aligns molecular structures with large-scale, noisy texts, extracting consensus information across comprehensive views. The second stage incorporates high-quality, structured knowledge from knowledge graphs. Drawing inspiration from knowledge-enhanced pre-training \cite{wang2021kepler, wang2022language}, we treat relations as specific types of views and describe them with texts. With contrastive and generative objectives, MV-Mol is endowed with the ability to grasp complementary information within different views.


We show the superior performance of MV-Mol through fine-tuning across various downstream tasks. Benefiting from view-based molecular representations and two-stage pre-training, MV-Mol achieves an average of 1.24\% absolute gains over the state-of-the-art method Uni-Mol \cite{zhou2023uni} on molecule property prediction \cite{wu2018moleculenet}. Additionally, MV-Mol exhibits a deeper understanding of connections between molecular structures and texts. On cross-modal retrieval, MV-Mol improves the top-1 retrieval accuracy by 12.9\% on average over the best-performing baselines. On cross-modal generation \cite{edwards2022translation}, MV-Mol also yields more accurate predictions, as validated by qualitative and quantitative studies.

Our contributions are summarized as follows: (1) To the best of our knowledge, MV-Mol is the first work that addresses the multi-view problem in molecular representation learning. (2) We propose to incorporate view information explicitly by jointly comprehending molecular structures and view prompts within a fusion network. (3) We develop a two-stage training pipeline to harvest multi-view molecular expertise from structured and unstructured knowledge sources that vary in quality and quantity.
\begin{figure*}[htpb]
\centering
    \includegraphics[width=\linewidth]{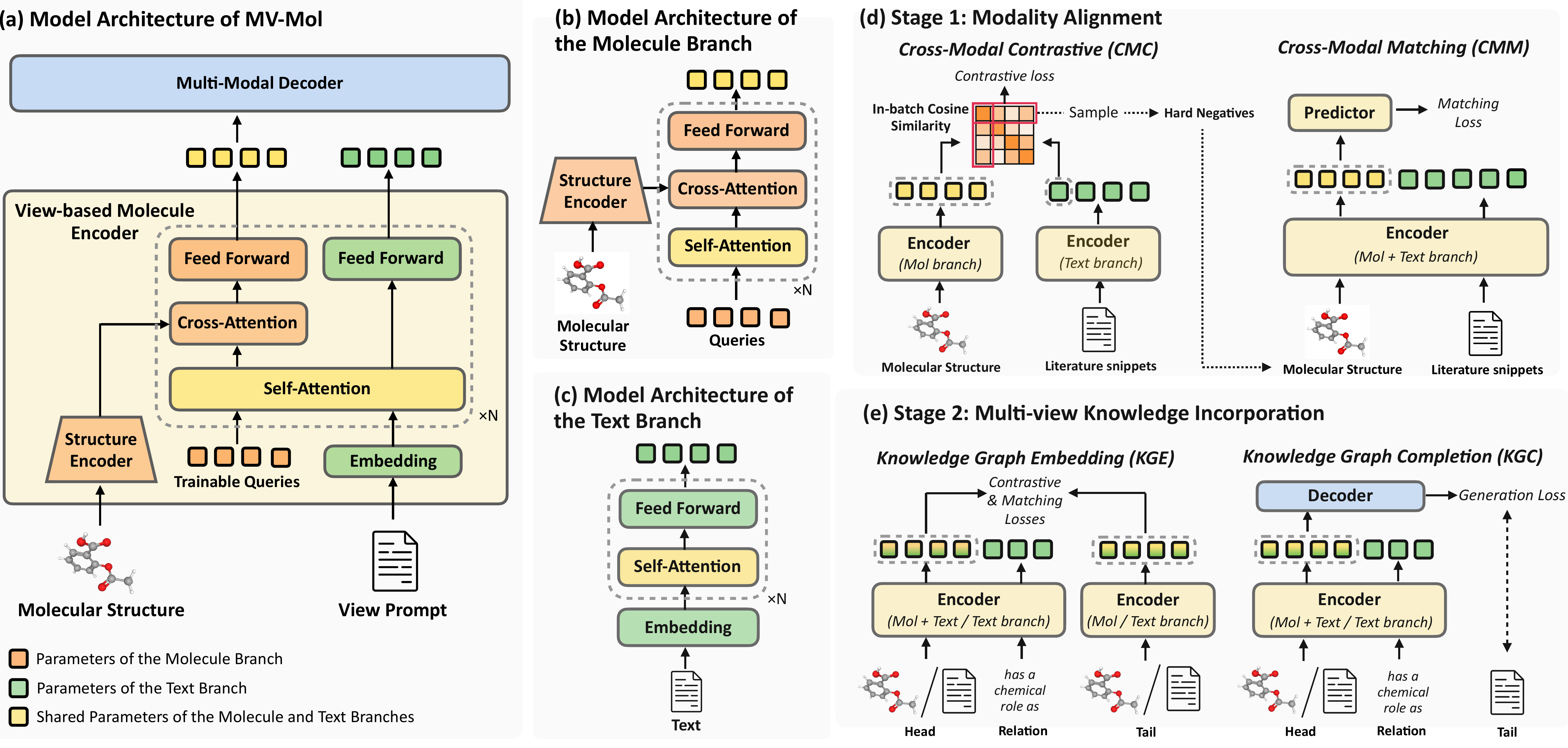}
\captionsetup{font={small,stretch=0.95}}
\caption{\textbf{Model architecture and pre-training pipeline of MV-Mol. (a)} MV-Mol is composed of a view-based molecule encoder and a multi-modal decoder. \textbf{(b)} The molecule branch of the view-based molecule encoder. \textbf{(c)} The text branch of the view-based molecule encoder. \textbf{(d)} We perform cross-modal contrastive and cross-modal matching for modality alignment. \textbf{(c)} We model relation as a textual prompt that constrains molecular knowledge from a specific view, and design knowledge graph embedding and knowledge graph completion objectives for multi-view knowledge incorporation. Both branches are activated when the head entity refers to a molecule.}
\label{fig:pipeline}
\end{figure*}
\section{Related work}
\subsection{Multi-view Molecular Representation Learning} 
Multi-view representation learning \cite{li2018survey, zheng2023comprehensive} aims to pursue a unified and comprehensive feature representation based on multi-view data. Most MRL models treat different forms of molecular structures as distinct views and model complementary and consensus information with different pre-training objectives. For example, DVMP \cite{zhu2023dual} treats 1D SMILES sequences and 2D molecular graphs as dual views. GraphMVP \cite{liu2021pre} treats 2D topologies and 3D geometries as two views, and proposes independent objectives for intra-view and inter-view modeling. MORN \cite{ma2022morn} treats the sequential, topological, and spatial information as different views and implements a two-stage adaptive learning strategy with multi-view fusion for molecular property prediction. \citep{wan2023molecules} treats atoms, scaffolds, and functional groups as independent views and proposes a prompt-based aggregation module to extract task-specific molecular representations. In this work, we extend the multi-view concept of molecules as their inherent properties and patterns within a certain context, which is defined flexibly by human-understandable textual prompts.

\subsection{Molecular Representation Learning with Heterogeneous Sources} 
Recently, MRL approaches that jointly harvest heterogeneous sources including molecular structures, biomedical texts, and knowledge graphs have surged. Drawing parallels from vision-language pre-training (VLP) \cite{radford2021learning, li2023blip, sun2023generative}, these models can be categorized as follows: (1) Generative models including KV-PLM \cite{zeng2022deep}, MolT5 \cite{edwards2022translation}, MolXPT \cite{liu2023molxpt}, Text+Chem \cite{christofidellis2023unifying}, GIMLET \cite{zhao2024gimlet}, BioMedGPT \cite{luo2023biomedgpt}, Mol-Instructions \cite{fang2023mol}, BioT5 \cite{pei2023biot5} and BioT5+ \cite{pei2024biot5+}. These models treat molecules and texts as two languages and jointly process them within a unified language model \cite{kenton2019bert, raffel2020exploring, radford2018improving}. (2) Contrastive models like MoMu \cite{su2022molecular}, MoleculeSTM \cite{liu2023multi}, CLAMP \cite{seidl2023enhancing}, KANO \cite{fang2023knowledge} and GODE \cite{jiang2023bi}. These models incorporate texts or knowledge graphs as cross-modal supervision signals and augment molecular representations with contrastive learning \cite{radford2021learning}. (3) Hybrid models including MolFM \cite{luo2023molfm}, MolCA \cite{liu2023molca}, GIT-Mol \cite{liu2023git} and 3D-MoLM \cite{li2024towards}. These models perform fine-grained feature integration with multi-modal fusion networks and incorporate both generative and contrastive objectives for pre-training. While MV-Mol also adopts a fusion network architecture, we distinguish it from prior works by our explicit modeling of views by jointly comprehending molecular structures and texts and our pre-training process that addresses data imbalance of heterogeneous knowledge sources.

\section{Method}
As depicted in Figure \ref{fig:pipeline}, MV-Mol is designed to learn multi-view molecular representations from heterogeneous sources. The core of our approach lies in modeling view information with text prompts. 

The remainder of this section is organized as follows: First, we provide a roadmap for extracting multi-view molecular expertise within heterogeneous sources from a knowledge-enhanced pre-training perspective (Sec. \ref{sec:pretrain_data}). Then, we introduce the model architecture for MV-Mol (Sec. \ref{sec:model}) that extracts view-based molecular representations. To address the heterogeneity of pre-training data, we develop a two-stage pre-training paradigm, harvesting unstructured knowledge from biomedical documents through modality alignment (Sec. \ref{sec:mod_align}) and structured knowledge from knowledge bases with knowledge incorporation (Sec. \ref{sec:know_inc}).

\subsection{A Knowledge-enhanced Pre-training Perspective for Multi-view MRL\label{sec:pretrain_data}}
In this section, we interpret multi-view MRL from the perspective of knowledge-enhanced pre-training \cite{wang2021kepler, wang2022language, hu2023survey}, providing a roadmap for methodology design. Inspired by OntoProtein \cite{zhang2021ontoprotein}, we formulate the pre-training data as triplets $(h, r, t)$, where $h$ and $t$ are head and tail entities, and $r$ signifies the relation. We define two types of entities, $E_{\text{mol}}$ and $E_{\text{text}}$. $e^{(s)}\in E_{\text{mol}}$ is denoted as a molecule node with structural information. $e^{(t)}\in E_{\text{text}}$ is denoted as a text node formulated by a token sequence. $r$ is also formulated as text tokens that describe the relation. Depending on the types of the head and tail entities, we further categorize pre-training data as molecule-text triplets, molecule-molecule triplets, and text-text triplets.

Knowledge-enhanced pre-training methods typically maximize the following objective:
\begin{equation}
    \mathcal{L}=S(f(h, r), g(t)),
\end{equation}
where $f(h, r)$ is the relation-transformed representation of the head entity, $g(t)$ is the representation of the tail entity, and $S(\cdot,\cdot)$ is a scoring function that measures the alignment of the triplet. We observe for $h\in E_{\text{mol}}$, the view-based molecular representations have an equivalent form of $f(h,r)$. Besides, structured knowledge graphs are indispensable sources for multi-view MRL, as both views and relations can be conceptualized as textual refinements over specific aspects of molecules. Therefore, multi-view MRL comprises the following key design components:
\begin{itemize}
    \item The model architecture for $f(h, r)$, tasked with extracting different feature representations of $h$ in the context of $r$. As $h$ and $r$ may come from different modalities, the model should jointly comprehend molecular structures and texts.
    \item The formulation of the scoring function $S(\cdot,\cdot)$, intended to harvest rich supervision signals from heterogeneous sources. 
\end{itemize}

\subsection{Model Architecture for MV-Mol\label{sec:model}}
The model architecture of MV-Mol is depicted in Figure \ref{fig:pipeline}(a), comprising 2 parts: (1) A view-based molecule encoder that jointly comprehends molecular structures and texts. (2) A multi-modal decoder that generates human-understandable texts.

\noindent\textbf{View-based Molecule Encoder.} We define the structure of a molecule as $\mathcal{M}=(\mathcal{V}, \mathcal{E}, \mathcal{C})$ where $\mathcal{V}$ represents atoms, $\mathcal{E}$ represents bonds, and $\mathcal{C}\in \mathbb{R}^{|\mathcal{V}|\times 3}$ represents the 3D coordinates for each atom. We leverage Uni-Mol \cite{zhou2023uni}, a 15-layer Transformer \cite{xu2018powerful} pre-trained on 209M 3D conformations, to encode $\mathcal{M}$. The pre-trained molecular model mitigates the disproportionality between molecular structures and text data \cite{su2022molecular}. The structure encoder $f_{\text{mol}}$ transforms $\mathcal{M}$ into feature representations for each atom, which we denote as $z^{(a)}$.

In MV-Mol, we use text prompts to model different views of molecules, represented by a sequence of $L$ tokens $\mathcal{T}=[x_1,x_2,\cdots,x_L]$. Consequently, the view-based molecule encoder is expected to comprehend molecular structures and texts simultaneously. To achieve this, we leverage Querying Transformer (Q-Former) \cite{li2023blip}, a novel multi-modal fusion architecture from vision-language pre-training. The Q-Former consists of two branches, each consisting of 12 bidirectional Transformer layers initialized with KV-PLM* \cite{zeng2022deep}. The molecule branch takes $K$ trainable query vectors as input embeddings. The query features extract pivotal information from atomic representations $z^{(a)}$ through cross-attention within every other Transformer layer. The text branch digests $\mathcal{T}$ as its input. The self-attention layers across the two branches are shared, allowing the queries to grasp fine-grained semantics from text prompts. In this way, the molecular representations are propagated within each Q-Former layer to incorporate information from different views. 

Overall, the view-based molecule encoder $f_v$ can transform the molecular structure to a fixed number of features with the structure encoder and the molecule branch of Q-Former, which is illustrated in Fig. \ref{fig:pipeline}b:
\begin{equation}
\label{eq:mr}
z^{(s)}=\left[z_1^{(s)}, z_2^{(s)}, \cdots, z_K^{(s)}\right]=f_{v}(\mathcal{M}).
\end{equation}

Using Q-Former's text branch, it can also comprehend natural language as in Fig. \ref{fig:pipeline}c:
\begin{equation}
\label{eq:text}
    z^{(t)}=\left[z_1^{(t)}, z_2^{(t)}, \cdots, z_L^{(t)}\right]=f_{v}(\mathcal{T}),
\end{equation}

Most importantly, as shown in Figure \ref{fig:pipeline}a, it can grasp view-based molecular representations through multi-modal feature fusion of molecular structures and texts:
\begin{equation}
\label{eq:vbmr}
    z^{(s,t)}=\left[z_1^{(s,t)}, z_2^{(s,t)}, \cdots, z_K^{(s,t)}\right]=f_{v}(\mathcal{M}, \mathcal{T}).
\end{equation}

\noindent\textbf{Multi-modal Decoder.} The multi-modal decoder allows MV-Mol to interpret the view-based molecular representations with natural language. We employ BioT5 \cite{pei2023biot5}, a molecular language model with 12 Transformer encoder layers and 12 Transformer decoder layers. The model has undergone multi-task pre-training on SELFIES strings of molecular structures, FASTA sequences of proteins, and biomolecule-related texts. For MV-Mol, we leverage the decoder branch of BioT5 as the text decoder $f_{\text{dec}}$, which transforms the Q-Former outputs into a sequence of $N$ tokens $y=[y_1,y_2,\cdots,y_N]$ by casual generation:
\begin{equation}
    y_i=\text{argmax}\left\{P\left(y_i\big |y_{<i},z^{(\cdot)}\right)\right\},
\end{equation}
where $y_{<i}$ denotes preceding tokens, $z^{(\cdot)}$ refers to either $z^{(s)}$, $z^{(t)}$ or $z^{(s,t)}$, and the conditional probability is modeled by the text decoder $f_{\text{dec}}$.

\noindent \textbf{Discussions.} While our model shares a similar architecture with MolCA \cite{liu2023molca} and 3D-MoLM \cite{li2024towards}, we emphasize that MV-Mol extracts view-based molecular representations in Eq. \ref{eq:vbmr} with both branches of the Q-Former. In contrast, prior works are unaware of view information and encode molecules as in Eq. \ref{eq:mr}, where the text branch of the Q-Former is deactivated.

\subsection{Modality Alignment with Large-scale Unstructured Knowledge\label{sec:mod_align}}
In this stage, we aim to capture the consensus information within numerous views of molecules and cultivate MV-Mol's capability of jointly comprehending molecular structures and texts. This is accomplished by aligning the feature space between molecular structures $h\in E_{\text{mol}}$ and the relevant contexts $t\in E_{\text{text}}$ from biomedical literature. We omit the relation $r$ due to the unaffordable costs of annotating millions of noisy texts. As shown in Figure \ref{fig:pipeline}b, the pre-training objectives are:

\noindent\textbf{Cross-Modal Contrastive Loss.} This objective maximizes the mutual information between the structural and textual representations. We define structure-text similarity as follows:
\begin{equation}
\label{eq:sim}
    \text{sim}\left(z^{(s)}, z^{(t)}\right)=\max_{i=1,2,\cdots,K}\left\{f_{\text{proj}}\left(z_i^{(s)}\right)^T f_{\text{proj}}\left(z^{(t)}_1\right)\right\},
\end{equation}
where $f_{\text{proj}}(\cdot)$ is a projection network composed of a fully-connected layer and $l_2$ normalization, and $z^{(t)}_1$ denotes the textual representation of the \texttt{[CLS]} token.

The objective function is calculated as:
\begin{equation}
\label{equ:cmc}
\begin{aligned}
    \mathcal{L}_{\text{cmc}}=-\frac{1}{2}&\left[\log\frac{\exp(\text{sim}[f_v(h),f_v(t)]/\tau)}{\sum_{h'}\exp(\text{sim}[f_v(h'),f_v(t)]/\tau)}\right.\\
    +&\left.\log\frac{\exp(\text{sim}[f_v(h), f_v(t)]/\tau)}{\sum_{t'}\exp(\text{sim}[f_v(h),f_v(t')]/\tau)}\right],
\end{aligned}
\end{equation}
where $h'$ and $t'$ are molecules and texts within the same mini-batch, and $\tau$ is a temperature hyper-parameter.

\noindent\textbf{Cross-Modal Matching Loss.} This objective fosters the fine-grained comprehension of molecular structures and texts by predicting whether they correspond to the same molecule. Following \citep{li2021align, li2023blip}, we acquire hard negatives for each sample by obtaining one misaligned structure for the text and one misaligned text for the structure from the same mini-batch based on the similarity scores defined in Eq. \ref{eq:sim}. The objective is calculated as follows:
\begin{equation}
\label{eq:cmm}
    \mathcal{L}_{\text{cmm}}=H\left[y_{\text{cmm}}(h, t),\ f_{\text{cmm}}\left(f_{v}(h, t)\right)\right],
\end{equation}
where $f_{\text{cmm}}$ is a predictor composed of an average pooling and a fully-connected layer, $y_{\text{cmm}}(\cdot, \cdot)$ is the ground truth label indicating the alignment of the molecule-text pair, and $H(\cdot, \cdot)$ denotes cross-entropy.

The overall objective is the summation of the two loss functions:
\begin{equation}
    \mathcal{L}_1=\mathbb{E}_{(h,t)\sim \mathcal{D}_1}(\mathcal{L}_{\text{cmc}}+\mathcal{L}_{\text{cmc}}),
\end{equation}
where $\mathcal{D}_1$ is the pre-training dataset.

\subsection{Multi-view Knowledge Incorporation with Heterogeneous Triplets \label{sec:know_inc}}
In this stage, we aim to incorporate high-quality structured knowledge into MV-Mol and capture the multi-view characteristics of molecules within three different types of triplets. Motivated by \cite{wang2021kepler, wang2022language}, we introduce relation transformed representations of the head entity, $z^{(h,r)}$, calculated as follows:
\begin{equation}
    z^{(h,r)}=\begin{cases}f_{v}(h, r), \quad \ \ \ h\in E_{mol}\\
    f_{v}(h\oplus r), \quad h\in E_{text}\end{cases}
\end{equation}
where $\oplus$ denotes concatenation. By modeling relations as view prompts, we capture view-based molecular representations within the first term of the equation. We connect $z^{(h,r)}$ and $t$ with objectives shown in Figure \ref{fig:pipeline}c, which we detail as follows:

\noindent\textbf{Knowledge Graph Embedding Loss.} This objective captures the global features within knowledge bases by aligning $z^{(h,r)}$ and the representations of the tail entity. We first employ the following contrastive objective: 
\begin{equation}
\label{eq:kge_c}
\begin{aligned}
    \mathcal{L}_{\text{kge\_c}}=-\frac{1}{2}&\left[\log\frac{\exp(\text{sim}_{\text{tri}}[z^{(h,r)},f_{v}(t)]/\tau)}{\sum_{h'}\exp(\text{sim}_{\text{tri}}[z^{(h',r)},f_{v}(t)]/\tau)}\right.\\+&\left.\log\frac{\exp(\text{sim}_{\text{tri}}[z^{(h,r)},f_{v}(t)]/\tau)}{\sum_{t'}\exp(\text{sim}_{\text{tri}}[z^{(h,r)},f_{v}(t')]/\tau)}\right],
\end{aligned}
\end{equation}
where $h'$ and $t'$ are head and tail entities within the same batch, and $\text{sim}_{\text{tri}}(\cdot,\cdot)$ calculates the representation similarity, which we detail in Appendix \ref{sec:app_model}. This objective is a generalized form of Eq. \ref{equ:cmc} that incorporates relation information from heterogeneous triplets. 

Then, we incorporate a matching loss similar to Eq. \ref{eq:cmm} for molecule-text and text-text triplets with the same negative sampling strategy for the head and tail entity. We omit molecule-molecule triplets as $f_v$ cannot process two molecules simultaneously. The objective is calculated as follows:
\begin{equation}
\mathcal{L}_{\text{kge\_m}}=H\left[y_{\text{cmm}}(h, r, t),\ f_{\text{cmm}}\left(f_{v}(h, r\oplus t)\right)\right].
\end{equation}

\noindent\textbf{Knowledge Graph Completion Loss.} This objective elicits MV-Mol to 'talk' about molecular knowledge from a specific view. We implement the language model-based tail entity prediction task \cite{qin2021erica, liu2023gpt} by feeding the relation-transformed representations $z^{(h,r)}$ into the text decoder to generate descriptions for the tail entity. The objective is calculated for molecule-text and text-text triplets in an auto-regressive manner:
\begin{equation}
    \mathcal{L}_{\text{kgc}}=\sum_{x_i\in t}H\left[x_i,P\left(x_i|x_{<i},z^{(h,r)}\right)\right],
\end{equation}
where the conditional probability is modeled by the text decoder $f_{\text{dec}}$.

The overall objective is calculated as follows:
\begin{equation}
    \mathcal{L}_2=\mathbb{E}_{(h,r,t)\in \mathcal{D}_2}(\mathcal{L}_{\text{kge\_c}}+\mathcal{L}_{\text{kge\_m}}+\mathcal{L}_{\text{kgc}}),
\end{equation}
where $\mathcal{D}_2$ is the knowledge graph.
\begin{table*}[hbpt]
\captionsetup{font={small,stretch=1.}}
\caption{AUROC scores for molecular property prediction on MoleculeNet. The best results are marked in bold, and the second-best results are underlined. $\uparrow$: the higher the better. $\dag$: Our implementation. -: Not reported in the original paper. w/o view: without view prompts.}
\label{tab:mp}
\begin{tabular}{lccccccccc}
\toprule
Model  & BBBP $\uparrow$ & Tox21 $\uparrow$ & ToxCast $\uparrow$ & SIDER $\uparrow$ & ClinTox $\uparrow$ & MUV $\uparrow$  & HIV $\uparrow$  & BACE $\uparrow$ & Avg $\uparrow$  \\
\midrule
GraphCL \cite{you2020graph}      & 67.5$_{\pm 3.3}$ & 75.0$_{\pm 0.3}$  & 62.8$_{\pm 0.2}$    & 60.1$_{\pm 1.3}$  & 78.9$_{\pm 4.2}$    & 77.1$_{\pm 1.0}$ & 75.0$_{\pm 0.4}$ & 68.7$_{\pm 7.8}$ & 70.64 \\
GraphMVP \cite{liu2021pre}     & 72.4$_{\pm 1.6}$ & 74.4$_{\pm 0.2}$  & 63.1$_{\pm 0.4}$    & 63.9$_{\pm 1.2}$  & 77.5$_{\pm 4.2}$    & 75.0$_{\pm 1.0}$ & 77.0$_{\pm 1.2}$ & 81.2$_{\pm 0.9}$ & 73.07 \\
GEM \cite{fang2022geometry} & 72.4$_{\pm 0.4}$ & 78.1$_{\pm 0.1}$ & 69.2$_{\pm 0.4}$ & \underline{67.2}$_{\pm 0.4}$ & 90.1$_{\pm 1.3}$ & \underline{81.7}$_{\pm 0.5}$ & 80.6$_{\pm 0.9}$ & 85.6$_{\pm 1.1}$ & 78.11 \\
Uni-Mol \cite{zhou2023uni} & 72.9$_{\pm 0.6}$ & \underline{79.6}$_{\pm 0.5}$ & \underline{69.6}$_{\pm 0.1}$ & 65.9$_{\pm 1.3}$ & 91.9$_{\pm 1.8}$ & \textbf{82.1}$_{\pm 1.3}$ & \underline{80.8}$_{\pm 0.3}$ & 85.7$_{\pm 0.2}$ & \underline{78.56} \\
\midrule
KV-PLM \cite{zeng2022deep}       & 66.9$_{\pm 1.1}$ & 64.7$_{\pm 1.8}$ & 58.6$_{\pm 0.4}$ & 55.3$_{\pm 0.8}$ & 84.3$_{\pm 1.5}$ & 60.2$_{\pm 2.9}$ & 68.8$_{\pm 4.9}$ & 71.9$_{\pm 2.1}$ & 66.29     \\
MoMu \cite{su2022molecular}         & 70.5$_{\pm 2.0}$ & 75.6$_{\pm 0.3}$  & 63.4$_{\pm 0.5}$    & 60.5$_{\pm 0.9}$  & 79.9$_{\pm 4.1}$  & 70.5$_{\pm 1.4}$ & 75.9$_{\pm 0.8}$ & 76.7$_{\pm 2.1}$ & 71.63 \\
MoleculeSTM \cite{liu2023multi} & 70.0$_{\pm 0.5}$ & 77.0$_{\pm 0.4}$  & 65.1$_{\pm 0.3}$    & 61.0$_{\pm 1.0}$  & \underline{92.5}$_{\pm 1.1}$  & 73.4$_{\pm 2.9}$ & 76.9$_{\pm 1.8}$ & 80.8$_{\pm 1.3}$ & 74.57 \\
MolCA \cite{liu2023molca} & 70.0$_{\pm 0.5}$ & 77.2$_{\pm 0.5}$ & 64.5$_{\pm 0.8}$ & 63.0$_{\pm 1.7}$ & 89.5$_{\pm 0.7}$ & - & - & 79.8$_{\pm 0.5}$ & - \\
GIT-Mol \cite{liu2023git} & 73.9$_{\pm 0.6}$ & 75.9$_{\pm 0.5}$ & 66.8$_{\pm 0.5}$ & 63.4$_{\pm 0.8}$ & 88.3$_{\pm 1.2}$ & - & - & 81.1$_{\pm 1.5}$ & - \\
BioT5$^{\dag}$ \cite{pei2023biot5} & \textbf{75.2}$_{\pm 0.1}$ & 75.5$_{\pm 0.3}$  & 64.4$_{\pm 0.1}$  & 62.4$_{\pm 0.1}$  & 85.8$_{\pm 0.6}$  & 75.9$_{\pm 0.7}$ & 79.7$_{\pm 0.4}$ & \textbf{89.2}$_{\pm 0.1}$ & 76.01 \\
\midrule
MV-Mol (w/o view)  & 73.0$_{\pm 0.2}$ & 79.7$_{\pm 0.1}$ & 69.7$_{\pm 0.2}$ & 65.3$_{\pm 0.2}$ & 93.2$_{\pm 0.2}$ & 81.6$_{\pm 0.4}$ & 79.9$_{\pm 0.5}$ & 87.3$_{\pm 0.2}$ & 78.71          \\
MV-Mol   & \underline{73.6$_{\pm 0.2}$} & \textbf{80.3$_{\pm 0.6}$}  & \textbf{70.0$_{\pm 0.4}$}       & \textbf{67.3$_{\pm 0.0}$}   & \textbf{95.6$_{\pm 1.6}$}  & \textbf{82.1$_{\pm 0.5}$}       & \textbf{81.4$_{\pm 0.3}$}   & \underline{88.2$_{\pm 0.4}$}  & \textbf{79.80}          \\
\bottomrule
\end{tabular}
\end{table*} 

\begin{table}[htpb]\small
\captionsetup{font={small,stretch=1.}}
\caption{Statistics of benchmark datasets. We present the number of molecules, data samples, prediction tasks, and split protocols.}
\label{tab:datasets}
\setlength\tabcolsep{2pt}
\begin{tabular}{llllc}
\toprule
Dataset  & \# Molecules & \# Samples & \# Tasks & Split                             \\
\midrule
BBBP     & 2,039        & 2,039      & 1        & \multirow{8}{*}{Scaffold (8/1/1)} \\
Tox21    & 7,831        & 7,831      & 12       &                                   \\
ToxCast  & 8,597        & 8,597      & 617      &                                   \\
SIDER    & 1,427        & 1,427      & 27       &                                   \\
ClinTox  & 1,478        & 1,478      & 2        &                                   \\
MUV      & 93.807       & 93,807     & 17       &                                   \\
HIV      & 41,127       & 41,127     & 1        &                                   \\
BACE     & 1,513        & 1,513      & 1        &                                   \\
\midrule
PCdes    & 11,112       & 11,112     & 1        & \multirow{2}{*}{Scaffold (7/1/2)} \\
MVST     & 7,102        & 16,996     & 1        &                                   \\
\midrule
CheBI-20 & 33,008       & 33,008     & 1        & Random (8/1/1)         \\
\bottomrule
\end{tabular}
\vspace{-0.1cm}
\end{table}

\section{Experiments \label{sec:exp}}
In this section, we conduct extensive experiments to demonstrate the effectiveness of the proposed model by answering the following questions:

\begin{itemize}
    \item \textbf{Q1.} Compared to prior MRL models, does MV-Mol provide better molecular representations?
    \item \textbf{Q2.} Does MV-Mol address the heterogeneity of molecular structures and texts?
    \item \textbf{Q3.} Do view-based molecular representations capture the consensus and complementary knowledge from different views? How do view prompts affect the representations?
    \item \textbf{Q4.} How do the design components within the two pre-training stages affect MV-Mol?
\end{itemize}

\subsection{Pre-training Setup}
\textbf{Pre-training Dataset.} For the modality alignment stage, we collect large-scale molecule-text pairs by named entity recognition (NER) \cite{nadeau2007survey} and entity linking \cite{kolitsas2018end} on 3.5M scientific publications following previous works \cite{zeng2022deep, liu2023molxpt, pei2023biot5}, and obtain a total of 60K molecules and 12M molecule-text pairs. For the knowledge incorporation stage, we construct a knowledge graph by combining public databases including CheBI \cite{degtyarenko2007chebi}, PubChem \cite{kim2016pubchem}, and DrugBank \cite{wishart2018drugbank}, which comprises 273K entities and 643K relations. During pre-training data collection, we exclude molecules within the test set of downstream datasets to alleviate information leakage. More details are presented in Appendix \ref{sec:app_pt_data}.

\noindent \textbf{Training Configurations.} MV-Mol comprises a molecular structure encoder with 47.3M parameters, a Q-Former with 180.8M parameters, and a text decoder with 140.2M parameters. We pre-train MV-Mol for 70K steps with a batch size of 256 for the modality alignment stage and 50K steps with a batch size of 192 for the knowledge incorporation stage. The pre-training procedure is performed on 4 NVIDIA A100 GPUs for 5 days. We use the AdamW \cite{loshchilov2018decoupled} optimizer with a weight decay of $5\times 10^{-2}$. The learning rate is linearly warmed up to $10^{-4}$ in the first 2K steps and then decreases to $10^{-5}$ following a cosine annealing strategy. The temperature $\tau$ for contrastive learning is fixed as $0.1$.

\subsection{Molecular Property Prediction (Q1)\label{sec:molprop}}
To assess the quality of molecular representations, we perform fine-tuning experiments on molecular property prediction, a widely adopted task in MRL.

\noindent \textbf{Datasets.} We adopt 8 classification datasets from MoleculeNet \cite{wu2018moleculenet}, a popular benchmark that covers diverse properties of molecules. In line with prior works \cite{liu2021pre, zhou2023uni}, we adopt Scaffold split with a train/validation/test ratio of 8/1/1. Statistics of the benchmark datasets are presented in Table \ref{tab:datasets}.

\noindent \textbf{Baselines.} We adopt two types of baseline MRL models:
\begin{itemize}
    \item Uni-modal MRL baselines: GraphCL \cite{you2020graph}, GraphMVP \cite{liu2021pre}, GEM \cite{fang2022geometry} and UniMol \cite{zhou2023uni}. These models are pre-trained solely with molecular structures including 2D topologies, 3D conformations, or both.
    \item Multi-modal MRL baselines: KV-PLM \cite{zeng2022deep}, MoMu \cite{su2022molecular}, Mole-culeSTM \cite{liu2023multi}, MolCA \cite{liu2023molca}, GIT-Mol \cite{liu2023git}, and BioT5 \cite{pei2023biot5}. These models incorporate heterogeneous pre-training data including molecular structures and texts.
\end{itemize} 

\noindent \textbf{Implementation Details.} We manually write descriptions for each dataset and feed the prompt and molecular structure into MVMol's encoder to obtain view-based molecular representations. Then, we feed the encoder outputs into a max-pooling layer and a projector for binary classification. The projector is composed of two fully connected layers with ReLU activation. We also implement a variant of our model named MV-Mol (w/o view) by removing view prompts and solely encoding molecules as in Eq. \ref{eq:mr}. Notably, we apply the prediction head with BioT5's encoder for a fair comparison of the representation quality. We perform experiments three times with different random seeds and report AUROC scores. Refer to Appendix \ref{sec:app_setup} for more details of our hyper-parameters and prompts.

\noindent \textbf{Results and Analysis.} The overall results of molecular property prediction are displayed in Table \ref{tab:mp}. The key observations are as follows: (1) Compared with the state-of-the-art model Uni-Mol, MV-Mol achieves an absolute gain of 1.24\% on average. Overall, MV-Mol performs best on 5 of 8 datasets and second-best on the remaining 3 datasets. This outstanding performance validates the informativeness of MV-Mol's molecular representations. (2) On SIDER, ClinTox, and BACE which consist of limited training samples, MV-Mol attains more significant improvements over Uni-Mol (1.4\% on SIDER, 3.7\% on ClinTox, and 2.5\% on BACE), which indicates that adapting molecular representations with view prompts brings more significant benefits to low-data scenarios. (3) Removing view descriptions leads to a decline of 0.79\% on average. This performance drop corroborates our claims that downstream applications require molecular knowledge from different views, and a universal molecular representation fails to capture the specific context of each task.

\subsection{Zero-shot Cross-modal Retrieval (Q2) \label{sec:cross_modality}}
\begin{table}[tpb]\small
\captionsetup{font={small,stretch=0.9}}
\caption{Zero cross-modal retrieval results on the test split of PCdes. The best results are marked in bold, and the second-best results are underlined. $\uparrow$: the higher the better.}
\label{tab:retrieval_pcdes}
\setlength\tabcolsep{2.5pt}
\centering
\begin{tabular}{llcccccccc}
\toprule
SubTask & Model & MRR $\uparrow$  & R@1 $\uparrow$ & R@5 $\uparrow$ & R@10 $\uparrow$ \\ \midrule
\multirow{7}{*}{S-T} & CLAMP \cite{seidl2023enhancing}              & 0.58 & 0.18    & 0.49    & 0.67         \\
& MoMu \cite{su2022molecular}                   & 11.10 & 5.93    & 15.20  & 20.46     \\
& MoleculeSTM \cite{liu2023multi}             & \underline{40.30} & \underline{26.99}    & \underline{55.33}    & \textbf{66.48}       \\
& MolCA \cite{liu2023molca} & 33.65 & 23.21 & 45.52 & 54.29 \\
& 3D-MoLM \cite{li2024towards} & 1.68 & 0.53 & 1.61 & 2.57 \\
& MV-Mol (w/o view) & 41.17 & 30.85 & 53.12 & 60.23 \\
& MV-Mol              & \textbf{45.25} & \textbf{36.66}    & \textbf{55.78}  & \underline{60.41}     \\
\midrule
\multirow{7}{*}{T-S} & CLAMP \cite{seidl2023enhancing}           & 0.43 & 0.08    & 0.27    & 0.54     \\
& MoMu \cite{su2022molecular}                   & 11.58 & 6.29    & 15.78  & 21.86     \\
& MoleculeSTM \cite{liu2023multi}             & 30.18 & 17.45    & 43.94    & \underline{57.71}     \\
& MolCA \cite{liu2023molca} & \underline{33.48} & \underline{23.70} & \underline{44.30} & 53.30 \\
& 3D-MoLM \cite{li2024towards} & 2.69 & 0.76 & 3.60 & 5.89 \\
& MV-Mol (w/o view) & 45.51 & 35.89 & 56.68 & 62.70 \\
& MV-Mol              & \textbf{46.54} & \textbf{37.24}    & \textbf{57.17}    & \textbf{62.97}     \\
\bottomrule
\end{tabular}
\end{table}

\begin{table}[t]\small
\captionsetup{font={small,stretch=0.9}}
\caption{Zero cross-modal retrieval results on the test split of MVST. The best results are marked in bold, and the second-best results are underlined. $\uparrow$: the higher the better.}
\label{tab:retrieval_mvst}
\setlength\tabcolsep{2.5pt}
\centering
\begin{tabular}{llcccccccc}
\toprule
SubTask & Model & MRR $\uparrow$  & R@1 $\uparrow$ & R@5 $\uparrow$ & R@10 $\uparrow$ \\  \midrule
\multirow{7}{*}{S-T} & CLAMP \cite{seidl2023enhancing}             & 0.34 & 0.07    & 0.23    & 0.30     \\
& MoMu \cite{su2022molecular}                   & 13.59 & 6.97    & 20.06  & 26.69      \\
& MoleculeSTM \cite{liu2023multi}             & 17.80 & 9.61    & \underline{25.98}    & \underline{35.59}     \\
& MolCA \cite{liu2023molca} & \underline{18.89} & \underline{14.25} & 24.43 & 27.33 \\
& 3D-MoLM \cite{li2024towards} & 1.09 & 0.19 & 1.32 & 2.22 \\
& MV-Mol (w/o view) & 24.87 & 16.40 & 34.08 & 41.32 \\
& MV-Mol              & \textbf{35.38} & \textbf{24.28}    & \textbf{47.92}  & \textbf{57.27}     \\
\midrule
\multirow{7}{*}{T-S} & CLAMP \cite{seidl2023enhancing}             & 0.33 & 0.04    & 0.23    & 0.53     \\
& MoMu \cite{su2022molecular}                   & 11.81 & \underline{5.16}    & 17.60  & 25.26     \\
& MoleculeSTM \cite{liu2023multi}              & \underline{12.68} & 4.86    & \underline{19.41}    & \underline{29.67} \\
& MolCA \cite{liu2023molca} & 11.05 & 4.86 & 18.13 & 22.43 \\
& 3D-MoLM \cite{li2024towards} & 1.44 & 0.26 & 2.00 & 3.09 \\
& MV-Mol (w/o view) & 20.86 & 9.35 & 33.37 & 44.26 \\
& MV-Mol              & \textbf{34.07} & \textbf{22.85}    & \textbf{45.73}    & \textbf{56.56}     \\
\bottomrule
\end{tabular}
\end{table}

We investigate whether MV-Mol captures rich molecular knowledge from heterogeneous sources including texts and knowledge graphs. Hence, we evaluate MV-Mol on cross-modal retrieval, which contains two sub-tasks: structure-to-text (S-T) retrieval and text-to-structure (T-S) retrieval. The former aims to retrieve the most relevant text that describes a given molecule, and the latter aims to retrieve a molecule that best fits the textual description.

\noindent \textbf{Datasets.} We incorporate two datasets: PCdes \cite{zeng2022deep} and MVST. The PCdes dataset is collected from the CheBI database and comprises 15K molecules with their biochemical definitions and property descriptions. MVST (Multi-View Structure-Text) is a novel dataset introduced by this work. It is collected from chemical, physical, and pharmacokinetic definitions in PubChem. It consists of 7.1K molecules, each corresponding with 2 or more texts from different views. Both datasets are partitioned by Scaffold split with a train/validation/test ratio of 7/1/2. Differing from prior works \cite{zeng2022deep, su2022molecular}, we perform retrieval on the whole test set and report MRR (mean reversed rank) and Recall at 1/5/10. More details of the datasets are presented in Appendix \ref{sec:app_dataset}.

\noindent \textbf{Baselines.} We compare MV-Mol with models that have undergone contrastive learning on molecule-text data including CLAMP \cite{seidl2023enhancing}, MoMu \cite{su2022molecular}, MoleculeSTM \cite{liu2023multi}, MolCA \cite{liu2023molca} and 3D-MoLM \cite{li2024towards}.

\noindent \textbf{Implementation Details.} On PCdes, we fix the view description as 'biochemical properties and functions'. On MVST, we write descriptions for physical, chemical, and pharmacokinetic views to generate different representations for the same molecule. To further improve the retrieval performance, we modify the re-ranking algorithm in \cite{li2021align} with an ensemble technique. We first retrieve the top-$k$ candidates based on structure-text similarity in Eq. \ref{eq:sim}. Then, we calculate the CMM logits in Eq. \ref{eq:cmm} for these $k$ candidates. Finally, we re-rank them by a linear combination of the structure-text similarity scores and the CMM logits.

\noindent \textbf{Results and Analysis.} The retrieval results on PCDes are presented in Tab. \ref{tab:retrieval_pcdes}. We observe that: (1) MV-Mol outperforms state-of-the-art methods by 4.95\% and 13.06\% in MRR for S-T and T-S retrieval. While MoleculeSTM achieves the best results in R@10 on S-T retrieval, it is worth noting that the model is pre-trained on texts similar to those in PCDes and molecules within the test set, raising information leakage concerns. Nevertheless, MV-Mol demonstrates effectiveness in bridging molecular structures and texts. (2) Adding the view description yields a notable gain of 3.08\% and 1.03\% in MRR for S-T and T-S retrieval, indicating that the view-based molecular representations align better with texts in PCdes.

Table \ref{tab:retrieval_mvst} shows the results on MVST, where MV-Mol yields substantial improvements, surpassing MoleculeSTM and MolCA by 16.49\% and 21.39\% in MRR for S-T and T-S retrieval. The greater performance gain suggests that MV-Mol is endowed with molecular knowledge from a wider range of views. Notably, removing view descriptions leads to a more significant decrease, which we attribute to the multi-view property of the MVST dataset. While a molecule may correspond with multiple descriptions from distinct views, MV-Mol could distinguish between them based on the view-based molecular representations.

\subsection{Cross-modal Generation (Q2)\label{sec:molcap}}
\begin{table}[tpb]\small
\setlength\tabcolsep{1pt}
\centering
\captionsetup{font={small,stretch=0.9}}
\caption{Molecule captioning results on the test split of ChEBI-20. 'BL' is short for 'BLEU', and 'R' is short for 'ROUGE'. The best results are marked in bold, and the second-best results are underlined. $\uparrow$: the higher the better. -: Not reported in the original paper.}
\label{tab:molcap}
\begin{tabular}{lccccccc}
\toprule
Model       & \#Params & BL-2 $\uparrow$ & BL-4 $\uparrow$ & R-1 $\uparrow$ & R-2 $\uparrow$ & R-L $\uparrow$ & METEOR $\uparrow$  \\
\midrule
MolReGPT \cite{li2023empowering}    & -         & 0.565  & 0.482  & 0.623   & 0.450   & 0.543   & 0.585     \\
MolT5-base \cite{edwards2022translation}  & 250M      & 0.540  & 0.457  & 0.634   & 0.485   & 0.578   & 0.569     \\
MolT5-large \cite{edwards2022translation} & 770M      & 0.594  & 0.508  & 0.654   & 0.510   & 0.594   & 0.612     \\
MoMu \cite{su2022molecular}       & 770M      & 0.599  & 0.515  & -       & -       & 0.593   & 0.597     \\
MolXPT \cite{liu2023molxpt}     & 350M      & 0.594  & 0.505  & 0.660   & 0.511   & 0.597   & 0.626      \\
Text+Chem \cite{christofidellis2023unifying}  & 250M      & 0.625  & 0.542  & 0.682   & 0.543   & 0.622   & 0.648          \\
MolCA \cite{liu2023molca} & 110M & 0.620 & 0.531 & 0.681 & 0.537 & 0.618 & 0.651 \\  
ChatMol \cite{zeng2023interactive}    & 220M      & 0.620  & 0.535  & 0.677   & 0.538   & 0.617   & 0.644     \\
GIT-Mol \cite{liu2023git} & 320M & 0.352 & 0.263 & 0.575 & 0.485 & 0.560 & 0.533 \\ 
BioT5 \cite{pei2023biot5}      & 252M      & \underline{0.635}  & \underline{0.556}  & \underline{0.692}   & \underline{0.559}   & \underline{0.633}   & \underline{0.656}      \\
\midrule
MV-Mol (w/o view)       & 452M      & 0.650  & 0.572  & 0.698   & 0.567   & 0.640   & 0.669 \\
MV-Mol       & 452M      & \textbf{0.653}  & \textbf{0.575}  & \textbf{0.699}   & \textbf{0.569}   & \textbf{0.640}   & \textbf{0.669}     \\
\bottomrule
\end{tabular}
\end{table}

Cross-modal generation also involves two sub-tasks: structure-to-text generation (molecule captioning) and text-to-structure generation (text-based molecule generation).

\noindent \textbf{Datasets.} We conduct experiments on ChEBI-20 \cite{edwards2021text2mol}, a widely-adopted dataset for both tasks. We follow the original split with a train/valid/test ratio of 8/1/1 and adopt evaluation metrics in \cite{edwards2022translation}. 

\noindent \textbf{Baselines.} We compare MV-Mol with the following baselines:
\begin{itemize}
    \item MolReGPT \cite{li2023empowering}. This baseline leverages a kNN few-shot strategy \cite{nie2022improving} for in-context learning \cite{dong2022survey} with ChatGPT \cite{chatgpt}. 
    \item Molecular language model baselines, including MolT5 \cite{edwards2022translation}, MolXPT \cite{liu2023molxpt}, Text+Chem \cite{christofidellis2023unifying}, ChatMol \cite{zeng2023interactive} and BioT5 \cite{pei2023biot5} that process 1D SMILES or SELFIES strings of molecules and texts within a unified language model.
    \item Fusion model baselines, namely MoMu \cite{su2022molecular}. The baseline connects the GNN outputs with MolT5 to solve molecule captioning. 
\end{itemize}

\noindent \textbf{Implementation Details.} For molecule captioning, we fix the view description as 'biochemical properties and functions'. In line with \citep{su2022molecular, liu2023git}, we concatenate the outputs of the MV-Mol encoder with the SELFIES representations of the BioT5 encoder and feed the results into our multi-modal decoder to generate the caption. For text-based molecule generation, we feed the textual representations in Eq. \ref{eq:text} into the MV-Mol decoder to generate the SELFIES string of the molecule.

\noindent \textbf{Results and Analysis.} Table \ref{tab:molcap} shows the results of molecule captioning. Compared to the state-of-the-art model BioT5, MV-Mol shows 1.8\% improvements in BLEU scores \cite{papineni2002bleu} and 1.3\% improvements in METEOR \cite{banerjee2005meteor} scores, indicating that it generates smoother and more semantically related molecular descriptions.

The performance comparison for text-based molecule generation is reported in Table 5. MV-Mol achieves state-of-the-art performance in most evaluation metrics, highlighted by an outstanding exact ratio of 0.438, surpassing state-of-the-art by 2.5\%. While MV-Mol underperforms Text+Chem in fingerprint similarity, we argue that these metrics are only calculated on valid molecules, which may yield over-optimistic results. 

Above all, the experiments show that MV-Mol addresses data heterogeneity and can flexibly translate between molecular structures and natural language.

\subsection{A Deeper Analysis of View-based Molecular Representations (Q3)}
\begin{figure}[tpb]
\centering
    \includegraphics[width=\linewidth]{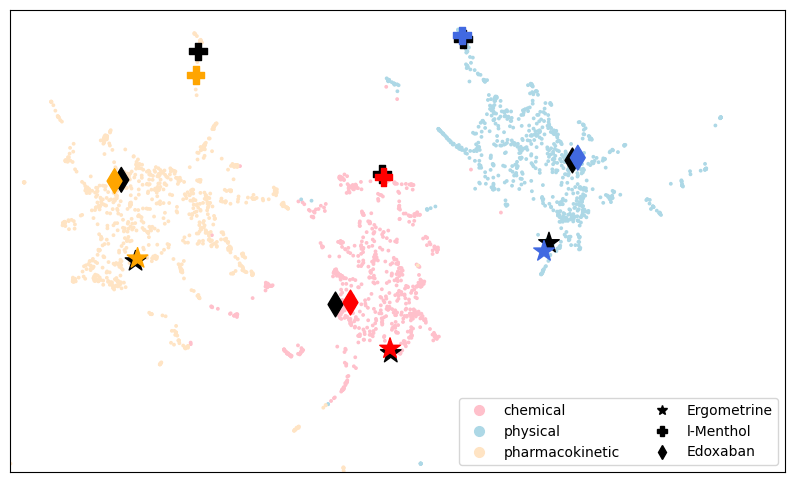}
    \captionsetup{font={small,stretch=0.9}}
\caption{Visualization of view-based molecular representations. We show the molecular representations from chemical, physical, and pharmacokinetic views. We also highlight three molecules and the representations of their textual descriptions from each view.}
\label{fig:vis_feats}
\end{figure}

\noindent \textbf{Visualization of View-based Molecular Representations.} To investigate how view descriptions affect molecular representations, we visualize the molecular features of MV-Mol on the MVST dataset with UMAP \cite{mcinnes2018umap}. As shown in Figure \ref{fig:vis_feats}, the molecular representations based on chemical, physical, and pharmacokinetic views exhibit clear separation along the x-axis. Furthermore, we highlight three representative molecules (indicated by colored shapes) along with corresponding texts from each view (indicated by black shapes), which are closely positioned. Notably, the partial order of the three molecules along the y-axis remains consistent under different views. These observations support our intuition that view-based molecular representations can concurrently capture the consensus and complementary molecular knowledge from different views.

\noindent \textbf{View Prompt Engineering.} Motivated by recent advances in prompt engineering \cite{chen2023unleashing, white2023prompt}, we explore how different view prompts affect the representations of our model. We perform analysis on BBBP, SIDER, ClinTox, and BACE from molecule property prediction, and implement the following strategies:
\begin{itemize}
    \item Empty prompt. This strategy refers to MV-Mol (w/o view), where the view information is not provided.
    \item Word prompt. We use the name of the dataset as the prompt.
    \item Sentence prompt. We manually write a brief definition of the prediction task as the prompt.
    \item Paragraph prompt. We write a well-rounded introduction to the dataset.
\end{itemize}

\begin{figure*}[htpb]
\begin{subfigure}[h]{0.62\linewidth}\small
\setlength\tabcolsep{2pt}
\centering
\captionsetup[table]{font={small,stretch=0.9}}
\caption{Table 5: Text-based molecule generation results on the test split of ChEBI-20. The best results are marked in bold, and the second-best results are underlined. $\uparrow$: The higher the better. $\downarrow$: The lower the better. -: Not reported in the original paper.}
\label{tab:t2mgen}
\begin{tabular}{lccccccccc}
\toprule
Model     & \# Params                     & BLEU $\uparrow$  & Exact $\uparrow$ & Valid $\uparrow$ & Levenshtein $\downarrow$ & \makecell[l]{MACCS\\ FTS $\uparrow$} & \makecell[l]{RDKit\\ FTS $\uparrow$} & \makecell[l]{Morgan\\ FTS $\uparrow$} & \\
\midrule
MolReGPT \cite{li2023empowering} & - & 0.790 & 0.139 & 0.887 & 24.910 & 0.847 & 0.708 & 0.624  \\
MolT5-base \cite{edwards2022translation} & 250M   & 0.779 & 0.082 & 0.786      & 25.188  & 0.787 & 0.661  & 0.601             \\
MolT5-large \cite{edwards2022translation} & 770M   & 0.854 & 0.311 & 0.905      & 16.071  & 0.834 & 0.746  & 0.684             \\
MoMu \cite{su2022molecular} & 235M & 0.815 & 0.183 & 0.863     & 20.520           & 0.847         & 0.737       & 0.678                  \\
MolXPT \cite{liu2023molxpt} & 350M & - & 0.215 & \underline{0.983} & - & 0.859 & 0.757 & 0.667  \\
Text+Chem \cite{christofidellis2023unifying} & 250M & 0.853 & 0.322 & 0.943 & 16.870 & \textbf{0.901} & \textbf{0.816} & \textbf{0.757} \\
ChatMol \cite{zeng2023interactive} & 220M & 0.848 & 0.258 & 0.947 & 16.759 & 0.883 & 0.790 & 0.726 \\
GIT-Mol \cite{liu2023git} & 250M & 0.756 & 0.051 & 0.928 & 26.315 & 0.738 & 0.582 & 0.519 \\
BioT5 \cite{pei2023biot5} & 252M & \underline{0.854} & \underline{0.413} & \textbf{1.000} & \underline{15.200} & 0.886 & 0.801 & 0.734 \\
\midrule
MV-Mol  & 252M  & \textbf{0.858} & \textbf{0.438} & \textbf{1.000} & \textbf{14.952}      & \underline{0.890}     & \underline{0.810}   & \underline{0.745}   \\
\bottomrule
\end{tabular}
\end{subfigure}
\hfill
\begin{subfigure}[h]{0.35\linewidth}
\centering
\includegraphics[width=\linewidth]{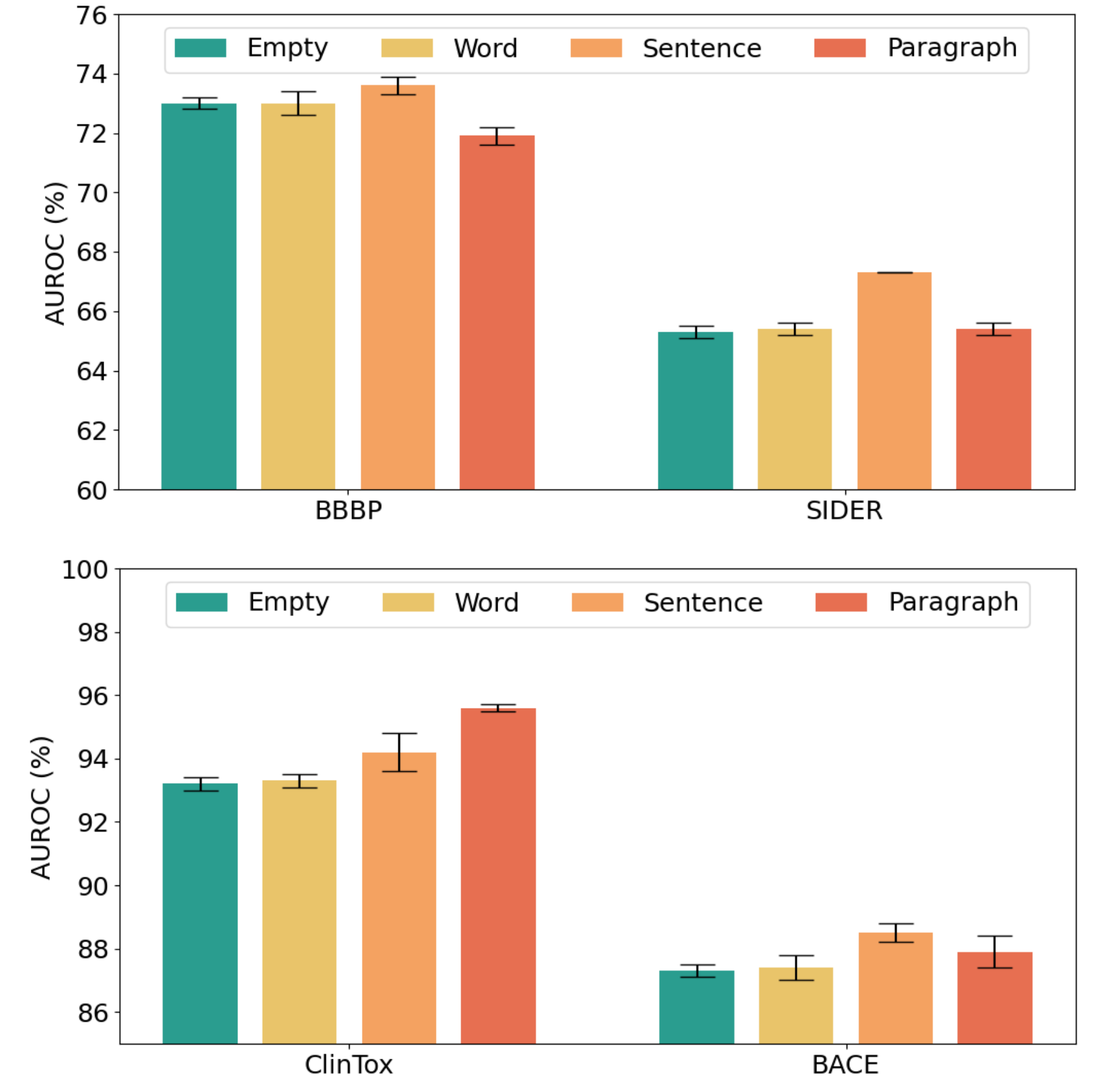}
\captionsetup{font={small,stretch=0.95}}
\caption{Figure 4: Experiment results on BBBP, SIDER, ClinTox, and BACE for different prompts.}
\label{fig:exp_prompt}
\end{subfigure}
\end{figure*}

As shown in Figure \ref{fig:exp_prompt}, the sentence prompt works best with BBBP, SIDER, and BACE, while the paragraph prompt yields the best results on ClinTox. We posit that BBBP, SIDER, and BACE focus on specific properties including blood-brain barrier penetration, adverse drug reactions, and target inhibition, where a single sentence suffices to encapsulate all relevant information. In ClinTox, however, more texts are necessary to describe clinical trial criteria and toxicity measures. Besides, the word prompt brings little improvement over the empty prompt. We attribute this to the abbreviated dataset names, which are not frequently used in the pre-training corpora and provide little meaningful information for MV-Mol.

\captionsetup[subfigure]{labelformat=empty}

\subsection{Ablation Studies (Q4)\label{sec:ablation}}
To explore the contribution of our two-stage pre-training, we compare the performances of MV-Mol by removing different pre-training objectives for uni-modal and multi-modal tasks and display the results in Table \ref{tab:ablation}. We observe that: (1) Knowledge graph completion (KGC) loss mainly contributes to cross-modal generation but compromises MV-Mol's performance on cross-modal retrieval (line 2). (2) Removing the knowledge incorporation stage leads to significant deterioration on all downstream tasks (line 3), highlighting the significance of incorporating multi-view knowledge of molecules. (3) The cross-modal matching objective also benefits all tasks (lines 3 and 4), as it cultivates the joint comprehension of molecules and texts. (4) Removing the modality alignment stage leads to moderate performance drops (line 5), indicating the indispensability of molecular knowledge from large-scale publications. (5) Compared with Stage 1 pre-training, Stage 2 brings more contributions, especially for retrieval tasks, which we attribute to two reasons. First, the knowledge incorporation stage is the key to multi-view MRL, allowing MV-Mol to jointly comprehend view prompts and molecular structures. Second, the knowledge graph enjoys higher quality than the noisy molecule-text pairs for the modality alignment stage.

\setcounter{table}{5}
\begin{table}[htpb]\small
\captionsetup{font={small,stretch=0.95}}
\caption{Comparison between MV-Mol with different pre-training objectives. For molecule property prediction, we report the average results on 8 datasets. For zero-shot cross-modal retrieval, we report the average results of Recall at 1/5/10 on the MVST dataset. For cross-modal generation, we report BLEU-2.}
\label{tab:ablation}
\centering
\begin{tabular}{ccccccccc}
\toprule
\multicolumn{2}{c}{Stage 1} & \multicolumn{2}{c}{Stage 2} & \multirow{2}{*}{Property} & \multicolumn{2}{c}{Retrieval} & Generation \\ 
 CMC & CMM & KGE & KGC  &                           & S-T           & T-S           & S-T                      \\
\midrule
\CheckmarkBold  & \CheckmarkBold & \CheckmarkBold  & \CheckmarkBold &  0.798  & 0.432  & 0.417   &    0.653        \\
\midrule
\CheckmarkBold & \CheckmarkBold & \CheckmarkBold & \XSolidBrush & 0.796   & 0.437  & 0.451   & 0.646      \\
\CheckmarkBold & \CheckmarkBold & \XSolidBrush & \XSolidBrush &   0.786  &   0.306   &  0.260   &  0.647     \\    
\CheckmarkBold & \XSolidBrush & \XSolidBrush & \XSolidBrush & 0.781    &   0.275   &  0.248   & 0.644       \\  
\XSolidBrush & \XSolidBrush & \CheckmarkBold & \CheckmarkBold &   0.787  &   0.395   &  0.400  & 0.650              \\
\bottomrule
\end{tabular}
\end{table}

\section{Limitations and broader impacts}
While our work presents promising results in molecular representation learning, several areas remain for future exploration: (1) Improving the scale and quality of the pre-training data for molecules from structured and unstructured knowledge sources. (2) Exploring the scaling laws of foundation models \cite{kaplan2020scaling} by incorporating large language models (LLMs) \cite{touvron2023llama, openai2023gpt4} into MV-Mol. (3) Applying MV-Mol to more biomedical entities such as proteins \cite{xu2023protst}, DNA and RNA sequences \cite{richard2024chatnt}, and cell transcriptomics \cite{zhao2024langcell}.

MV-Mol bears promise for accelerating biomedical research. However, there is a concern that MV-Mol may be misused to generate potentially dangerous or toxic molecules. Therefore, it is essential to ensure the responsible and ethical use of the model. We emphasize that MV-Mol should be employed solely for research purposes, and any potential medical applications of MV-Mol should undergo comprehensive experimental evaluations. 
\section{Conclusion}
In this paper, we present MV-Mol, a molecular representation learning model to harness multi-view molecular expertise from structured and unstructured knowledge sources. To capture the consensus and complementary information across different views, we propose to model views explicitly with textual prompts. We leverage a multi-modal fusion architecture to extract view-based molecular representations, and address the heterogeneity of structured and unstructured knowledge with varying quality and quantity with a two-stage pre-training strategy. Through extensive experiments, we demonstrate the superiority of MV-Mol on molecular property prediction and cross-modal translation. Under thorough analysis aimed at safety, MV-Mol has the potential to deliver unprecedented advancements to the biomedical research community.

\begin{acks}
This research is supported by the National Key R\&D Program of China (No. 2022YFF1203002) and PharMolix Inc.
\end{acks}

\bibliographystyle{ACM-Reference-Format}
\bibliography{main}

\appendix
\setcounter{figure}{0}
\setcounter{table}{0}
\setcounter{equation}{0}
\renewcommand{\thefigure}{A\arabic{figure}}
\renewcommand{\thetable}{A\arabic{table}}
\renewcommand{\theequation}{A\arabic{equation}}
\section{Details of MV-Mol \label{sec:app_model}}
The representation similarity $\text{sim}_{\text{tri}}(\cdot,\cdot)$ in Eq. \ref{eq:kge_c} is calculated as follows:
\begin{equation}
\begin{aligned}
    &\text{sim}_{\text{tri}}(z^{(h,r)},z^{(t)})\\&=\begin{cases}\max_{i=1,2,\cdots,K}\left\{f_{\text{proj}}(z^{(h,r)}_i)^Tf_{\text{proj}}(z^{(t)}_1)\right\},h\in E_{\text{mol}},t\in E_{\text{text}},\\
    \max_{i,j=1,2,\cdots,K}\left\{f_{\text{proj}}(z_i^{(h,r)})^T f_{\text{proj}}(z^{(t)}_j)\right\},h,t\in E_{\text{mol}},\\
    f_{\text{proj}}(z^{(h,r)}_1)^T f_{\text{proj}}(z^{(t)}_1),h,t\in E_{\text{text}}\\
    \end{cases}
\end{aligned}
\end{equation}
\section{Pre-training Data Collection\label{sec:app_pt_data}}
Following \cite{zeng2022deep, su2022molecular}, we perform named entity recognition (NER) \cite{nadeau2007survey} and entity linking (EL) \cite{kolitsas2018end} to obtain molecule-text pairs from biomedical literature. Specifically, we first sample 321K frequently accessed molecules from PubChem \cite{kim2016pubchem}. Then we identify mentions of these molecules within 3.5M publications from S2ORC \cite{lo2020s2orc}. We start with the corresponding sentence of the mention and randomly extend the context by the previous or next sentence within the paragraph until the text snippet exceeds 256 tokens. Consequently, we collect 12M molecule-text pairs for 60K molecules. 

We also collect a knowledge graph for molecules by combining public databases including PubChem \cite{kim2016pubchem}, CheBI \cite{degtyarenko2007chebi}, DrugBank \cite{wishart2018drugbank} and FORUM \cite{delmas2021building}. We first extract 196K molecules from these databases and perform deduplication. Then, we collect the connections between these molecules and other entities as follows:
\begin{itemize}
    \item We collect drug property descriptions that span diverse views from PubChem. We treat each description as a unique node and connect it with the corresponding molecule.
    \item We collect drug targets, drug enzymes, drug carriers, and drug transporters from DrugBank. We also collect BindingDB compounds and protein targets with $Ki\le 10 nM$.
    \item We collect drug-disease and drug-ontology connections from FORUM with $\text{q}\_\text{value}<10^{-6}$.
    \item  We collect drug-ontology and ontology-ontology connections from CheBI with libCheBI toolkit \cite{swainston2016libchebi}.
\end{itemize}

In total, the knowledge graph comprises 273K entities and 643K relationships. Table \ref{tab:kg_stat} presents the overall statistics of the knowledge graph. Following \cite{zhou2023uni}, we calculate 3D conformations by the MMFF algorithm \cite{tosco2014bringing} within the RDKit \cite{landrum2013rdkit} package for molecules. 

\begin{table}[tpb]
\centering
\setlength\tabcolsep{4pt}
\captionsetup{font={small,stretch=0.95}}
\caption{Statistics of the collected knowledge graph. The left column denotes the entity or relation type, and the right column denotes the number of entities and relations.}
\label{tab:kg_stat}
\begin{tabular}{ll}
\toprule
\multicolumn{2}{l}{Entities}  \\
\midrule
Molecules  & 196,454 \\
Targets & 403 \\
Ontologies & 22,076 \\
Diseases & 2,835 \\
Properties & 51,882 \\
All & 273,650 \\
\midrule
\multicolumn{2}{l}{Relations} \\
\midrule
Molecule-Target Interaction & 23,870 \\
Molecule-Ontology Connection & 252,555 \\
Molecule-Disease Association & 172,241 \\
Molecule-Physical Property & 6,166 \\
Molecule-Chemical Property & 38,206 \\
Molecule-Pharmacokinetic Property & 7,510 \\
Molecule-Molecule Connection & 119,793 \\
Ontology-Ontology Connection & 22,706 \\
All & 643,047 \\
\bottomrule
\end{tabular}
\vspace{0.2cm}
\end{table}

\balance
\section{Benchmark Datasets\label{sec:app_dataset}}
\subsection{Molecular Property Prediction}
We adopt 8 classification datasets for molecular property prediction, including BBBP (Blood-Brain Barrier Penetration), Tox21 (Toxicology in the 21st Century), ToxCast (Toxicity), SIDER (Side Effects Resource), ClinTox (Clinical Trial Outcomes and Toxicity), MUV (Maximum Unbiased Validation), HIV (HIV inhibition), and BACE (Human $\beta$-secretase 1 Binding).
\subsection{Cross-Modal Retrieval}
For cross-modal retrieval, we incorporate the following datasets:
\textbf{PCDes.} The original dataset \cite{zeng2022deep} comprises 15K molecules and their descriptions in CheBI. We filter out 3,888 molecules that have 
appeared in our pre-training dataset. Unlike the original experiment setting, we adopt the Scaffold split with a train/validation/test ratio of 7/1/2 and perform zero-shot retrieval on the test set with the entire paragraph. 

\noindent \textbf{MVST.} This dataset is introduced as part of our work. Different from PCDes, each molecule within MVST corresponds to multiple descriptive texts, distinguished by a given view. The dataset primarily focuses on 3 types of views:
\begin{itemize}
    \item The chemical view. The data for this view is directly derived from the CheBI description of molecules within the PubChem database.
    \item The physical view. We assemble multiple columns within PubChem including the general physical description, color, form, order, boiling point, melting point, flashing point, solubility, density, vapor pressure, stability (shelf life), and decomposition. The text is organized as '{\textit{COLUMN\_NAME}: \textit{COLUMN\_DESCRIPTION}}'. We randomly sample rows if the assembled text exceeds 256 tokens.
    \item The pharmacokinetic view. We also incorporate multiple columns from PubChem including drug indication, disposal methods, drug-food interactions, toxicity summary, metabolism, therapeutic uses, inhalation systems, absorption, distribution, exertion, biological half-life, adverse effects, and carcinogenicity. The organization of the text and length control follows the same as the physical view.
\end{itemize}

Following previous works \cite{edwards2022translation, zeng2022deep}, we replace the name of the molecule within texts with '\textit{the molecule}'. After collecting data from each view, we sample molecules with textual descriptions from 2 or more views and partition them by Scaffold split with a train/validation/test ratio of 7/1/2.

\subsection{Cross-Modal Generation}
We incorporate the CheBI-20 dataset, a widely adopted dataset collected from the CheBI database. We follow the original random split with a train/validation/test ratio of 8/1/1.

\section{Downstream Experiment Details\label{sec:app_setup}}
We manually write descriptions for each dataset in MoleculeNet as view prompts to improve molecular representations. Specifically:
\begin{itemize}
    \item BBBP. The prompt is \textit{'blood-brain barrier penetration (permeability)'}.
    \item Tox21. The prompt is \textit{'Qualitative toxicity measurements including nuclear receptors and stress response pathways'}.
    \item ToxCast. The prompt is \textit{'Qualitative toxicity measurements'}.
    \item SIDER. The prompt is \textit{'adverse drug reactions (ADR) for 27 system organ classes'}.
    \item ClinTox. The prompt is \textit{'Qualitative data of drugs if they failed clinical trials for toxicity reasons'}.
    \item MUV. The prompt is \textit{'Subset of PubChem BioAssay designed for validation of virtual screening techniques'}.
    \item HIV. The prompt is \textit{'Experimentally measured abilities to inhibit HIV replication'}.
    \item BACE. The prompt is \textit{'Binding results for human $\beta$-secretase 1 (BACE-1)'}.
\end{itemize}

\end{document}